**Title: Comparison of Object Detection Algorithms Using Video and Thermal Images Collected from a UAS Platform: An Application of Drones in Traffic Management**

**Hualong Tang**
Department of Civil and Environmental Engineering
University of South Florida, Tampa, FL 33620
Email: hualongtang@usf.edu

**Joseph Post**
Department of Civil and Environmental Engineering
University of South Florida, Tampa, FL 33620
Email: japost@usf.edu

**Achilleas Kourtellis**
Center for Urban Transportation Research
University of South Florida, Tampa, FL 33620
mail: kourtellis@cutr.usf.edu

**Brian Porter**
Professional Engineer
Florida Department of Transportation
Tampa, FL 33612
Email: brian.porter@dot.state.fl.us

**Yu Zhang**
Department of Civil and Environmental Engineering
University of South Florida, Tampa, FL 33620
Email: yuzhang@usf.edu

Word Count: 6,221 words + 4 table (250 words per table) = 7,221 words

*Tang, H., Post, J., Kourtellis, A., Porter, B., and Zhang, Y.*

## ABSTRACT

There is a rapid growth of applications of Unmanned Aerial Vehicles (UAVs) in traffic management, such as traffic surveillance, monitoring, and incident detection. However, the existing literature lacks solutions to real-time incident detection while addressing privacy issues in practice. This study explored real-time vehicle detection algorithms on both visual and infrared cameras and conducted experiments comparing their performance. Red Green Blue (RGB) videos and thermal images were collected from a UAS platform along highways in the Tampa, Florida, area. Experiments were designed to quantify the performance of a real-time background subtraction-based method in vehicle detection from a stationary camera on hovering UAVs under free-flow conditions. Several parameters were set in the experiments based on the geometry of the drone and sensor relative to the roadway. The results show that a background subtraction-based method can achieve good detection performance on RGB images (F1 scores around 0.9 for most cases), and a more varied performance is seen on thermal images with different azimuth angles. The results of these experiments will help inform the development of protocols, standards, and guidance for the use of drones to detect highway congestion and provide input for the development of incident detection algorithms.

*Tang, H., Post, J., Kourtellis, A., Porter, B., and Zhang, Y.*

## INTRODUCTION

Unmanned Aerial Vehicles (UAVs) are increasingly being used by state and local transportation agencies for a variety of purposes. Infrastructure inspection and disaster management (e.g., rockfall inspection, damage assessment, flood/ice jam monitoring, etc.) are the most common applications of UAVs by these agencies, but there is increasing interest in using the technology to monitor traffic in real time [1]. Traditionally, loop detectors, radar detectors, cameras and other conventional sensors are used for measuring and monitoring traffic conditions; however, these sensors are limited to fixed sites. Roadway incidents cause the reduction of road capacity and lead to significant congestion and vehicle delay if they are not addressed quickly. Automated detection of incidents with traditional sensors has been developed but is not applied widely. Road Rangers are used to continuously patrol the roadways, looking for collisions and stranded motorists and then responding to those incidents. Nevertheless, continuously patrolling roadways is both costly and manpower-intensive. UAVs offer the opportunity to rapidly and autonomously reconnoiter large sections of roadway, with little investment in fixed infrastructure. Using UAVs, sensors such as regular and infrared (thermal) cameras can collect traffic information, and real-time incident detection algorithms can be developed for prompt response to detected incidents.

The National Institute for Congestion Reduction (NICR) funded research at the University of South Florida (USF) and the University of Puerto Rico Mayaguez (UPRM) to investigate real-world applications of UAV video, focusing on traffic data collection and incident detection. The research is planned to proceed in two phases. For Phase I, the objective of the experiments was to select and refine image processing algorithms and quantify the performance of the overall system (including operator, drone, sensor, and processing algorithms) under varying settings. Phase II will focus on developing a real-time incident detection algorithm with selected sensors and explore how outcomes of the study can be integrated into daily operations of traffic management centers.

Phase I started with selecting appropriate UAVs and sensing technologies for the research, followed by experimental design and collection of sample video. The team flew two drones alongside interstate highways in the Tampa, Florida, area, collecting extensive video in both the visual and infrared bands. The geometry of the drone and sensor relative to the roadway was varied as part of the experimental design, as was drone speed, ambient lighting, and level of traffic congestion. Image processing algorithms were selected for analyzing the video data, and performance of UAV systems was compared with different performance metrics.

The results of these experiments will help inform the development of protocols, standards, and guidance for the use of drones to detect highway congestion and provide input for the development of an incident detection algorithm to be developed in Phase II.

This paper summarizes the experiments conducted and results obtained by the USF team during Phase I. The remainder of the paper is organized as follows: Section 2 summarizes the literature review; Section 3 describes the research approach, including experimental design, selection of image processing algorithms and performance metrics; Section 4 presents the filed data collection; Section 5 summarizes the results of the data analysis; and Section 6 concludes this phase of study.

## LITERATURE REVIEW
### Application of drones in traffic management

Drones have been used in many aspects of road traffic surveillance, monitoring, and analysis. One of the most studied areas is using drones to monitor traffic, such as incident monitoring and driver behavior monitoring. [2] used drones to investigate an incident site by broadcasting the site location to a traffic management center (TMC) and also reconstructing a virtual 3D model of the site for further analysis. [3] applied a quadcopter to monitor highway incidents. In their framework, the highway patrol deploys the quadcopter equipped with First Person View (FPV) to capture the incident video footage and transmit the footage to a ground station or TMC in real-time. [4] proposed to use UAS to detect and analyze abnormal driving behaviors to promote highway safety. Researchers have also studied methods of tracking vehicles from drones [5], analyzed the accuracy of position estimation of objects based on aerial imagery captured by drones [6], and studied video processing algorithms for tracking moving vehicles using drone video. Traffic analysis has been reported in several papers. Although data were collected by drones, the focus in those studies was primarily on the development of methods to analyze traffic data to obtain traffic parameters such as speed, flow, density, shockwaves, and queue length [7, 8].

**Comparison of RGB and thermal cameras and images**

Sensors are an important part of any data collection apparatus. UAVs already use a suite of sensors for flight control and navigation. They are often equipped with Global Positioning System (GPS), Inertial Navigation Sensors (INS), Micro-Electro-Mechanical Systems (MEMS) gyroscopes and accelerometers, and Altitude Sensors (AS) and one or more sensors for their primary task of video recording or other type of data collection [9]. UAVs can acquire very detailed information of observed objects by using a wide range of cameras such as Red Green Blue (RGB) sensor cameras, infrared, or thermal cameras that can be useful for getting aerial imaging for vehicles and tracking them [2].

The most widely used sensors for data collection with UAV are RGB cameras [11]. These cameras provide multiple features such as high-resolution images and video recording. Lightweight RGB cameras provide the flexibility of using them with UAVS where weight is an issue. RGB cameras and thermal cameras have been used in surveillance, agriculture, inspection filmmaking, and many other industries [10]. For use in traffic monitoring systems, which is the main focus of this paper, RGB cameras are useful as flight aids and as data collectors to detect and process scenes involving vehicles [12]. RGB cameras also have limitations; the quality of the RGB visual data is highly dependent on the light present at the scene—the more intense the surrounding light, the better the quality of the image.

A different visual sensor is a thermal camera, which measures the infrared radiation emitted by all objects. The energy of the radiation mainly depends on the object temperature. This can be useful when flying UAVs in different weather and light conditions. Compared to RGB cameras, thermal cameras have more flexibility capturing visual data with low level of lighting. Thermal cameras provide enough information for object detection by capturing lower resolution images with fewer details; this can be useful when privacy is important, as thermal images do not reveal vehicle license plates or the face of a person [13]. Using thermal cameras also helps reduce the pre-processing time of the frame before any real-time object detection and tracking algorithms, as the step of blurring part of the frames to obscure private information can be eliminated. Thermal cameras, therefore, have great potential for traffic monitoring purposes with the correctly calibrated algorithms in place.

**Vehicle detection algorithms**

To extract traffic information, vehicles on the road first need to be detected. Then, based on the detected results, traffic flow parameters can be estimated. Vehicle detection is often regarded as an application of object detection, which is a computer technology related to computer vision and image processing that has been studied abundantly. Object detection involves two main tasks—localization and classification. Localization is locating the object in the form of bounding box coordinates, and classification is predicting the object class. If all the vehicles are deemed a single class, the object detection needs to deal only with the task of localization. This problem can be treated as a moving object detection problem, assuming cars are moving on the ground. The methods applied in object detection and moving object detection are discussed below.

Methods for object detection can be categorized into neural network-based or non-neural network approaches. For non-neural network approaches, features such as histogram of oriented gradients (HOG), Haar, and scale-invariant feature transform (SIFT) are first computed and then fed into a classification model such as a support vector machine (SVM) to create classifiers [14-16]. After the rise of deep learning, neural network-based approaches were developed to provide a higher detection accuracy based on the convolutional neural network (CNN), among which there are two major categories—two-stage or region-based detector and single-stage detector. The region-based convolutional neural networks (R-CNNs) are models that involve a region proposal stage to extract a region of interest (ROI). R-CNN applies selective search algorithms to extract 2,000 region proposals, or ROIs, from the image. Each region proposal is then warped and fed into a convolutional neural network before the classification module and bounding box regressor are applied [17]. The major problems of R-CNN are that inference time is very slow and the training process is very complex, as it requires training of three separate modules. To improve the speed of R-CNN, the same author built a faster detector, named Fast R-CNN, by passing the input image directly to the CNN to extract features before region proposals in such a way that only one CNN needs to run, as opposed to running 2,000 CNNs over 2,000 proposals [18]. Faster R-CNN further improved speed by replacing the selective search algorithm with a small convolutional network—a region proposal network—and achieved an end-to-end deep learning network [19]. Faster R-CNN is hundreds of times faster than Fast R-CNN, with a frame rate of 5 FPS on a GPU, but it can perform real-time detection tasks.

In the last few years, new architectures (single stage) have been created to address the inference time limitation of the R-CNN family. The two state-of-the-art single-stage detectors are single-shot detector (SSD) and You Only Look Once (YOLO). Different from region-based detectors that first generate region proposals and then feed them to classification/regression heads, single-stage detectors use a single convolutional network to make predictions of the bounding boxes and the class probability in one shot. YOLO has evolved to version 4 (YOLOv4) with a real-time speed of over 65 FPS and 43.5 percent mean average precision (mAP) for the MS COCO dataset. SSD321 can achieve 28.0 mAP and over 16 FPS. Both YOLOv4 and SSD can outperform Faster R-CNN in mAP [20-22].

Some classic approaches to detect moving objects include consecutive frame difference, optical flow, and background subtraction [23, 24]. Consecutive frame difference methods suffer robustness issues, and optical flow methods usually require many calculations, making real-time detection challenging. Background subtraction methods achieve a good balance between robustness and real-time detection, making them the most popular method in the literature. Background subtraction methods extract the moving foreground from the background and output a binary mask that separates the foreground and background pixels. To extract the foreground, background models are first created. The main difference between different types of background subtraction models is how the background model is built. Researchers have proposed approaches based on statistical models, machine learning models, and signal processing models. Statistical models such as single Gaussian, Mixture of Gaussians (MOG), and Kernel Density Estimation have been widely used in the literature [25-27]. Representation learning (e.g., GRASTA, incPCP) and neural networks (e.g., CNNs) have also been applied to background modeling [28-30]. Based on signal processing, researchers used signal estimation models, transform domain functions, and sparse signal recovery models to model the background [31-33].

Background subtraction methods are designed for static cameras assuming the background is not moving. When the camera is moving, the background cannot be modeled correctly. However, learning-based methods such as machine learning and deep learning can detect from a moving camera. However, there are other approaches inspired by background subtraction to deal with a moving camera, such as panoramic background subtraction, dual cameras, motion compensation, motion segmentation, and multi planes; all of these were designed for motion detection but have difficulty detecting objects of very low speed [34-37].

## EXPERIMENTAL DESIGN AND PERFORMANCE METRICS

To inform operational concepts for the use of drones for traffic monitoring, this project explored the effectiveness of drone congestion detection as a function of the geometry of the drone and sensor relative to the highway. Flying at higher altitudes and viewing the roadway at oblique angles can make more of the roadway visible to the sensor and thus might increase the rate at which the drone can cover a road. On the other hand, the accuracy of the detection algorithms may be negatively affected. There also are many drone operating restrictions, some imposed by the Federal Aviation Administration (FAA) (e.g., altitude limits and operations over people) and others relating to obstacle avoidance (radio towers, power lines, overpasses, etc.) that can restrict the drone's viewing perspective.

**Figure 1** illustrates the geometry of a drone and sensor relative to a monitored highway and defines some of the experimental parameters. In this project, we varied the height above ground *DH*, depression of the sensor relative to the horizontal plane $\alpha$, and azimuth of the sensor relative to the roadway $\psi$. We collected both RGB and infrared imagery for free-flowing traffic condition with a stationary camera. **Table 1** presents the experimental design. For all these experiments, the horizontal distance between the drone and the roadway, or the roadway offset DDR, was fixed at 100 ft.

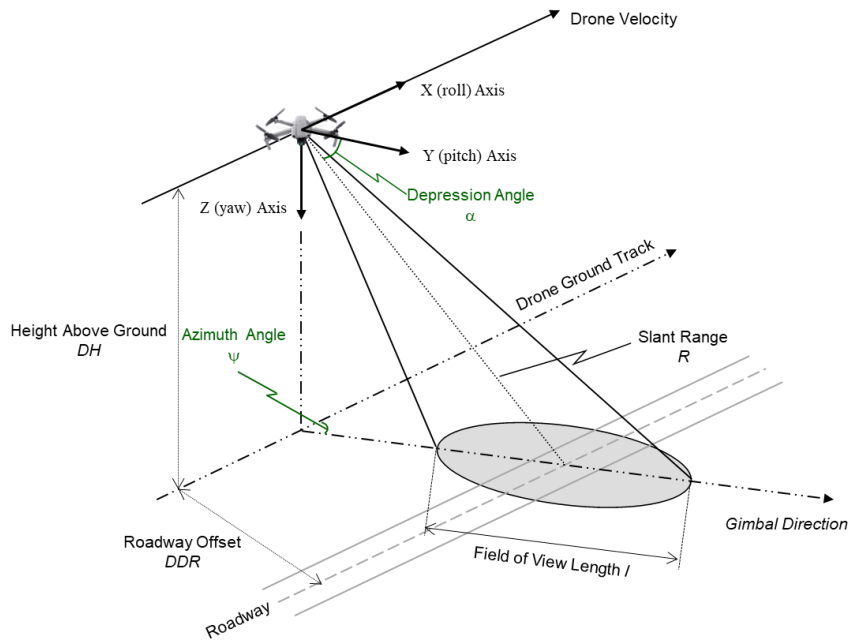

**Figure 1. Drone and sensor geometries**

**Table 1 Experimental Design**

| Traffic Intensity | Height Above Ground DH (ft) | Azimuth $\psi$ (deg) | Depression $\alpha$ (deg) |
|---|---|---|---|
| Free-Flowing (Without Congestion) or With Congestion | 50 | 45 | 45 |
| | | 90 | 70-90 |
| | | 135 | 45 |
| | 100 | 45 | 45 |
| | | 90 | 70-90 |
| | | 135 | 45 |
| | 200 | 45 | 45 |
| | | 90 | 70-90 |
| | | 135 | 45 |
| | 300 | 45 | 45 |
| | | 90 | 70-90 |
| | | 135 | 45 |
| | 400 | 45 | 45 |
| | | 90 | 70-90 |
| | | 135 | 45 |

In addition to varying the physical relationship between the drone, sensor, and roadway, the relationship between algorithm performance and slant range also were explored by virtually restricting the field of view of the sensor during image processing. It was expected that as slant range to the roadway increases, the performance of the vehicle detection algorithms will decrease. Therefore, experiments were conducted restricting the field of view length, $l$, when processing the imagery. In addition, data were for congested traffic condition and with a moving camera. We are currently analyzing these data and will share results with reviewers later.

**Selection of image processing algorithms**

Based on a survey of vehicle detection algorithms, the pros and cons of some representative models used in the literature are summarized in **Table 2**. First tested were algorithms on free-flow traffic, the focus of this paper. Although background subtraction-based methods cannot apply to moving cameras, their fast computation time and ease of implementation make them a desirable option for detecting moving vehicles in real-time when the camera is static. The algorithms will be tested on congested traffic and with a moving camera during ongoing work not included in this paper. Learning-based methods will be used, as they work well with low-speed objects and moving cameras. Given their higher accuracy and faster inference time, single-stage deep neural networks (YOLO v4) will be used to conduct detection for congested traffic and a moving camera.

**Table 2 Representative Methods and Their Pros and Cons in Vehicle Detection**

| Method | Background Subtraction-based Methods | | Machine Learning | Two-stage Deep Neural Networks | Single-stage Deep Neural Networks | |
|---|---|---|---|---|---|---|
| **Models/ algorithms** | Gaussian Mixture-based background/ foreground segmentation algorithm | Statistical background image estimation and per-pixel Bayesian segmentation | Cascade classifiers | R-CNN family | SSD | YOLO |
| **Camera** | Stationary | | Stationary/moving | | | |
| **Pros** | 1. Low computation time 2. Easy to implement | | 1. Training relatively short compared to deep learning 2. Low CPU power requirement | 1. High accuracy | 1. Trains faster than R-CNN 2. Real-time detection 3. Good balance between accuracy and speed | |
| **Cons** | 1. Not for moving or vibrating camera 2. Difficult to detect slow-moving objects 3. Many parameters to tune, often tricky | | 1. Object shape needs to be consistent 2. Change in rotation will affect performance | 1. High computation time 2. Large training time 3. GPU required | 1. GPU required 2. May be less accurate than R-CNN | |

**Figure 2** describes the steps applied to detect vehicles in free-flow traffic with a stationary camera using an approach based on the background subtraction method. A Gaussian Mixture-based Background/Foreground Segmentation Algorithm is used to create the background model [38]. Input frames are fed into the background model to separate the foreground (moving objects) from the static background. A binary image is output, with the foreground mask representing the moving objects. The binary image is further processed by two morphological transformations, opening and dilation. The opening removes noise by conducting erosion operation, followed by dilation operation. Closing (the reverse of opening, dilation followed by erosion) is applied after opening to remove small holes in the foreground objects. Contours are then generated based on the foreground masks. Noise is further filtered out by using a contour area threshold; any contours smaller than the threshold value are removed. Finally, the bounding boxes of detected vehicles are obtained according to the contour's coordinates.

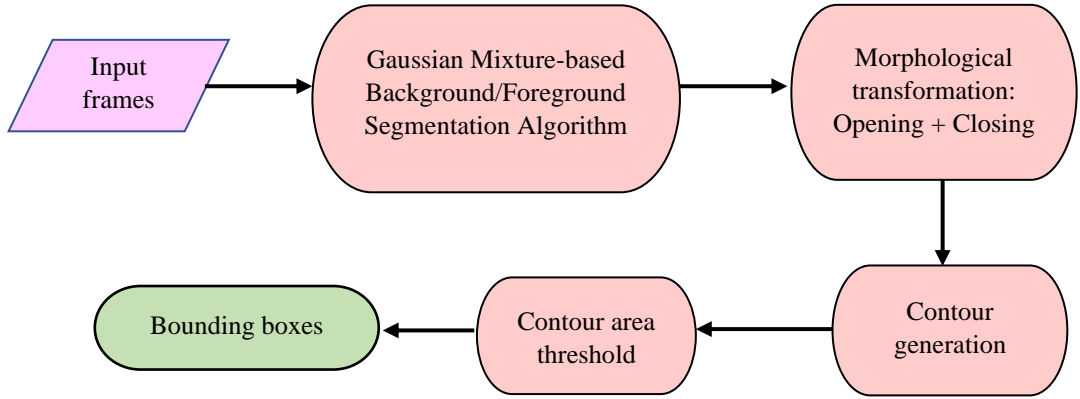

**Figure 2 Background-subtraction based approach.**

**Performance metrics**

To evaluate the performance of the image processing algorithms, performance metrics of precision and recall were calculated using the true positive (TP), false positive (FP), and false negative (FN) counts. The definition of precision and recall are as follows:

$$precision = \frac{TP}{TP+FP} \qquad (1)$$

$$recall = \frac{TP}{TP+FN} \qquad (2)$$

F1 score was also calculated, which is the harmonic mean of precision and recall, taking both metrics into account.

$$F1 = 2 \times \frac{precision \times recall}{precision + recall} \qquad (3)$$

**CASE STUDY SITE AND FIELD DATA COLLECTION**

**Data Collection Sites**

Determining road segments that would be suitable for data collection was the first step. Although the purpose of the study was to apply UAV video to identify non-recurrent congestion caused by incidents, it is difficult to capture such events with limited time and effort of data collection. Thus, first investigated were historical crash data to identify freeway sections with higher historical crash rates. Then, identified were small freeway segments in these sections where recurrent congestion is likely to occur during peak hours due to roadway capacity changes.

In addition to the crash and congestion analysis described above, the sites had to meet specific requirements for flying a UAS. As it was not desirable to go through the process of obtaining a Certificate of Authorization from the FAA, which can be time-consuming, test locations outside of controlled airspace were selected. **Figure 3** is an extract from the Visual Flight Rules (VFR) Terminal Area

Chart for Tampa. Tampa International Airport is the largest airport in the Tampa area occupying the most airspace; no operations could be conducted in its Class B airspace, which extends down to the surface, limiting test site locations.

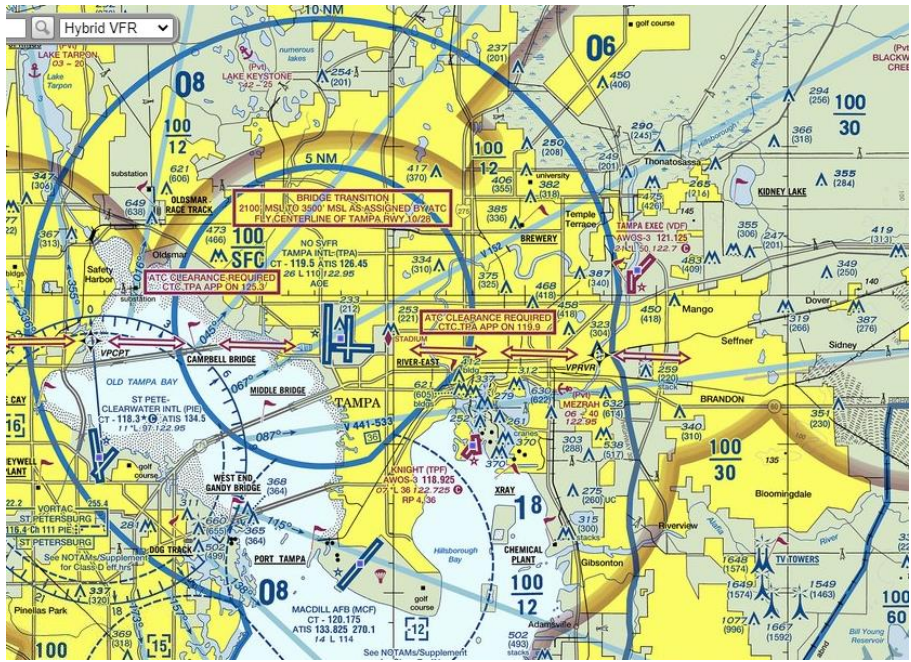

**Figure 3 VFR Terminal area chart of Tampa—used to ensure no airspace rules broken during data collection.**

The first road segment selected for data collection was along Interstate 75, as shown in **Figure 4**. The team used an empty parking lot of a nearby establishment to launch the UAS. The location was north of exit 279 on I-75, which location offered a clear space for a visual line of sight operations and proximity to the interstate lanes. This location was used for data collection under free-flowing speed conditions. Videos were recorded on Saturdays between 12:00–4:00 PM on clear days to ensure clear video quality. This location was not used for congested conditions.

The second segment, on I-275, was selected primarily for its high likelihood of congested conditions, as shown in **Figure 5**. The location was an empty parking lot adjacent to I-275 and next to an exit ramp that leads to E Bird St. Due to the location of the ramp, congestion was expected during PM peak hours on the right-side lanes. The road has an overpass and is elevated at the location due to the Hillsborough River and an arterial road underneath. Drone data showed an elevation difference of 30 ft from the parking lot to the surface of the interstate. This elevation was added to the height parameters so the height above the roadway was accurate.

The parking lot next to I-275 was used as a launched site for all data collection under congested conditions. A ramp between the parking lot and the interstate added distance to the roadway from the drone. Data collection occurred on three Fridays between 4:00–6:00 PM during peak hours.

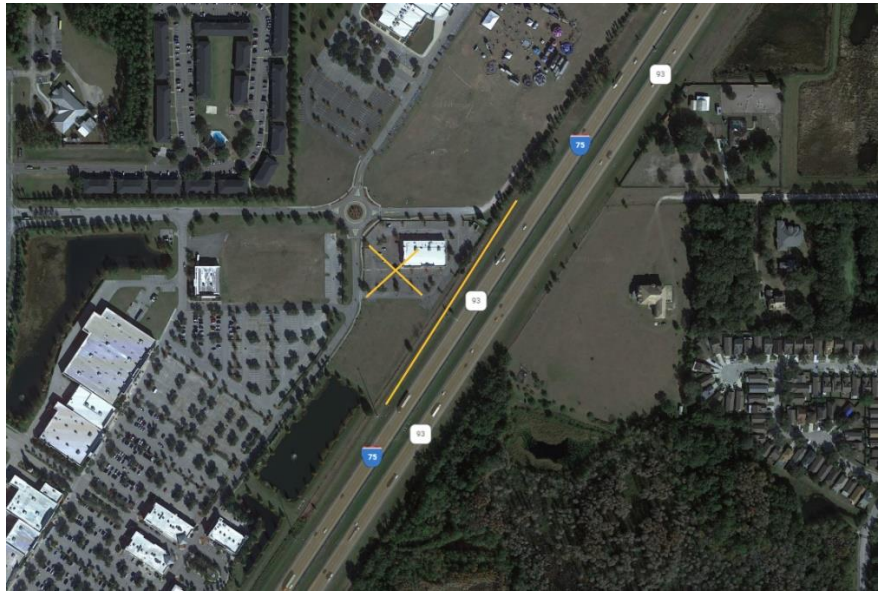

**Figure 4 Location of drone operations for uncongested data collection of I-75 traffic. "X" depicts location of drone operator and launch site; yellow line shows path of drone during collection at speed**.

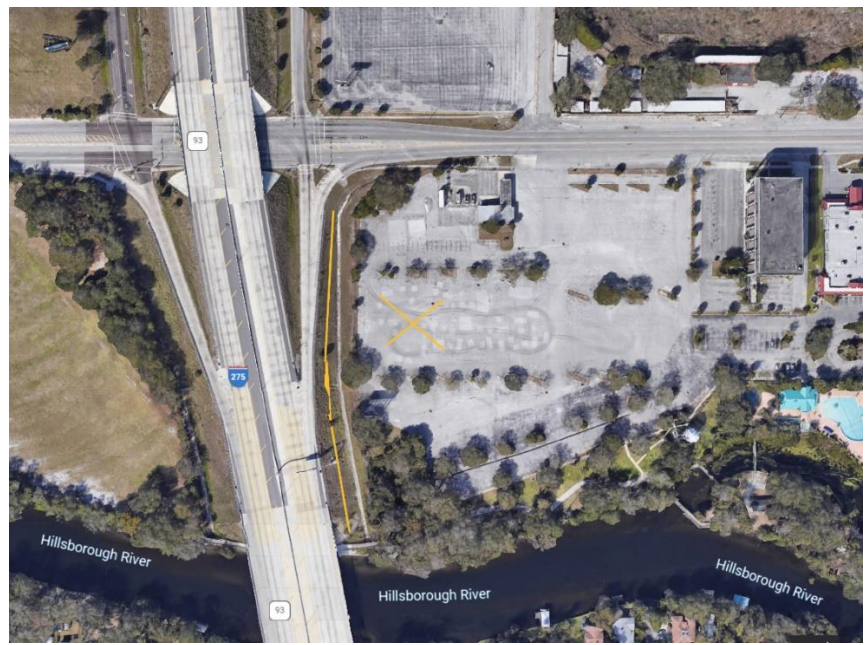

**Figure 5 Location of drone operations for congested conditions on I-275. "X" marks location of drone operator; yellow line shows path of drone during collection at speed.**

**Data Collection Parameters**

      To collect data needed for evaluation of the algorithms, the team created a set of parameters that would vary to establish comparison metrics for the different algorithms used. Each combination of parameters created a scenario, and video of traffic was recorded so the algorithms could be exercised and compared. **Figure 1** shows the different parameters considered for data collection. Not all parameters were varied, as this would create an unnecessarily large number of combinations that would complicate

data collection. **Table 3** shows the selected parameters varied for the scenarios. Video was captured at heights of 50, 100, 200, 300, and 400 ft; azimuth angles of 45, 90, and 135°; drone velocities of 0 and 5 mph; and elevation angles between 45 and 90°. All videos were collected during clear sky conditions and in the afternoon. The offset from the road was kept at 100 ft, and each video was recorded for 2 min. The combinations of these parameters resulted in 48 different scenarios under free-flow conditions and another 48 under congested conditions. Both RGB and thermal cameras were used for collecting video data for these scenarios.

**Table 3 Scenarios for Data Collection**

| Scenario No. | Height (h) | Azimuth (ψ) | Velocity (v) | Time of Day/Light Conditions | Depression Angle (α) | Offset (d) | Video Length (min) |
|---|---|---|---|---|---|---|---|
| 1 | 50ft, | 45 | | Afternoon – sunny | 45 | 100ft | |
| . | 100ft, 200ft, 300ft, | 90 | 0-20 mph | Afternoon – sunny | 45 | 100ft | 2 min each (96 min total) |
| 48 | 400ft | 135 | | Afternoon – sunny | 70–90 | 100ft | |

**OUTCOMES OF EXPERIMENTS**

This paper presents the experimental results of fixed station and free-flow traffic condition and compare the performance of RGB and thermal sensors with the background subtraction-based approach described in the experimental design.

When the drone was at a lower level (100 ft or 200 ft) with $\psi = 45°$ or $135°$, the field of view from the camera could go as far as to the horizon, which made the vehicles very far from the drone to capture in the frame. However, due to the small size of the far-away vehicles, the capability of the algorithm identifying the vehicles will be reduced. Thus, to restrict the field of view length, trail-and-error efforts were conducted; it was determined that the frame should be cut off 2/5 and detection conducted only on the bottom 3/5 of the frame. Traffic from both directions was detected, and a mask was used on the frame to cut off the frame and separate road directions. **Figure 6** shows an example of detections on images (azimuth angle $\psi = 135°$) from 50–400 ft after using a mask to cut off the frame and separate road directions.

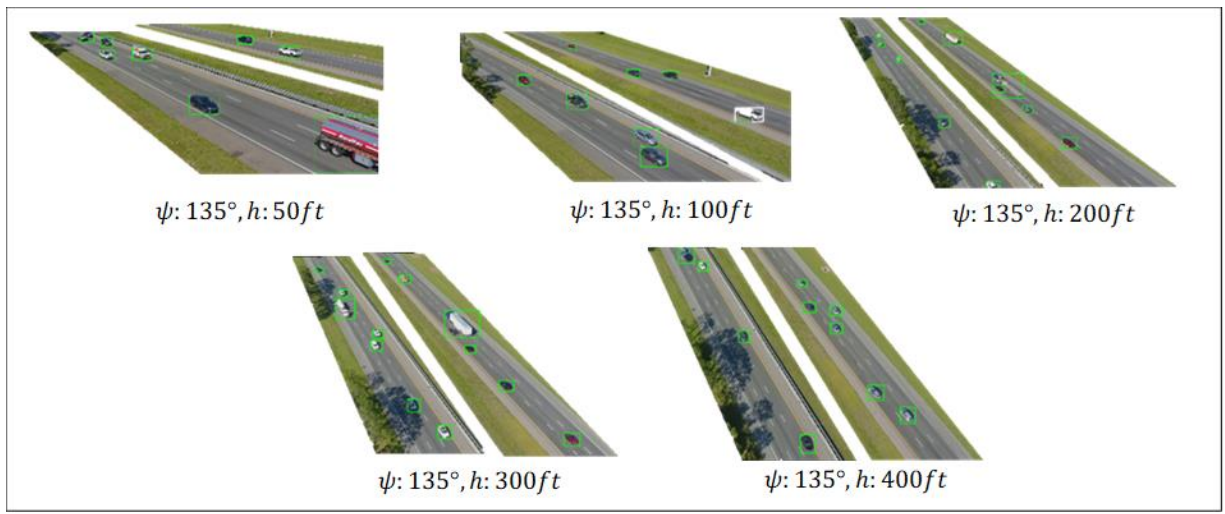

$\psi$: 135°, $h$: 50$ft$      $\psi$: 135°, $h$: 100$ft$      $\psi$: 135°, $h$: 200$ft$

$\psi$: 135°, $h$: 300$ft$      $\psi$: 135°, $h$: 400$ft$

**Figure 6 Detection on images (azimuth angle $\psi = 135°$) after using mask to cut off frame and separate road directions.**

Table shows the comparison of GRB and thermal videos of 15 experimental scenarios (five heights 50ft, 100ft, 200ft, 300ft, 400ft and three azimuth angles 45°, 90°, 135°). Detection results were collected on every 5th frame from the 200th to 700th frame for each video/scenario and performance metrics were computed. For example, for a video of height $h_1$ and azimuth angle $\psi_1$, we first manually counted ground truth (TP+FN) and TPs at each frame and collected the number of detected vehicles (TP+FN) from using a background subtraction algorithm. Precision and recall could then be calculated, as could an F1 score given precision and recall:

$$Precision_{h_1, \psi_1} = \frac{\Sigma_f TP_f}{\Sigma_f (TP_f + FP_f)} \quad f = \{200, 205, 210, \dots, 700\} \tag{4}$$

$$Recall_{h_1, \psi_1} = \frac{\Sigma_f TP_f}{\Sigma_f (TP_f + FN_f)} \quad f = \{200, 205, 210, \dots, 700\} \tag{5}$$

Statistics of the performance metrics for RGB and infrared images from different scenarios are shown in Table 4.

**Table 4 Results from RGB images of azimuth angle $\psi = 45°$ at different heights.**

*Results from RGB images*

| RGB | 50ft–45° | | 100ft–45° | | 200ft–45° | | 300ft–45° | | 400ft–45° | |
|---|---|---|---|---|---|---|---|---|---|---|
| Metrics | South | North | South | North | South | North | South | North | South | North |
| TP | 20 | 79 | 204 | 353 | 257 | 499 | 622 | 621 | 289 | 304 |
| FP | 5 | 7 | 22 | 34 | 49 | 24 | 31 | 14 | 20 | 26 |
| FN | 1 | 6 | 15 | 48 | 42 | 47 | 34 | 123 | 26 | 17 |
| Precision | 0.800 | 0.919 | 0.903 | 0.912 | 0.840 | 0.954 | 0.953 | 0.978 | 0.935 | 0.921 |
| Recall | 0.952 | 0.929 | 0.932 | 0.880 | 0.860 | 0.914 | 0.948 | 0.835 | 0.917 | 0.947 |
| F1 | 0.870 | 0.924 | 0.917 | 0.896 | 0.850 | 0.934 | 0.950 | 0.901 | 0.926 | 0.934 |
| | *50ft–90°* | | *100ft–90°* | | *200ft–90°* | | *300ft–90°* | | *400ft–90°* | |
| Metrics | South | North | South | North | South | North | South | North | South | North |
| TP | 145 | 131 | 294 | 100 | 110 | 136 | 273 | 232 | 457 | 581 |
| FP | 3 | 4 | 109 | 49 | 15 | 8 | 44 | 73 | 43 | 65 |
| FN | 8 | 37 | 72 | 20 | 26 | 10 | 12 | 25 | 26 | 36 |
| Precision | 0.980 | 0.970 | 0.730 | 0.671 | 0.880 | 0.944 | 0.861 | 0.761 | 0.914 | 0.899 |
| Recall | 0.948 | 0.780 | 0.803 | 0.833 | 0.809 | 0.932 | 0.958 | 0.903 | 0.946 | 0.942 |
| F1 | 0.963 | 0.865 | 0.765 | 0.743 | 0.843 | 0.938 | 0.907 | 0.826 | 0.930 | 0.920 |
| | *50ft–135°* | | *100ft–135°* | | *200ft–135°* | | *300ft–135°* | | *400ft–135°* | |
| Metrics | South | North | South | North | South | North | South | North | South | North |
| TP | 184 | 386 | 339 | 175 | 606 | 538 | 583 | 588 | 253 | 245 |
| FP | 63 | 74 | 79 | 74 | 53 | 83 | 45 | 56 | 0 | 34 |
| FN | 23 | 54 | 37 | 18 | 34 | 36 | 47 | 50 | 47 | 40 |
| Precision | 0.745 | 0.839 | 0.811 | 0.703 | 0.920 | 0.866 | 0.928 | 0.913 | 1.000 | 0.878 |
| Recall | 0.889 | 0.877 | 0.902 | 0.907 | 0.947 | 0.937 | 0.925 | 0.922 | 0.843 | 0.860 |
| F1 | 0.811 | 0.858 | 0.854 | 0.792 | 0.933 | 0.900 | 0.927 | 0.917 | 0.915 | 0.869 |

**Results from infrared images**

| IFR | 50ft–45° | | 100ft–45° | | 200ft–45° | | 300ft–45° | | 400ft–45° | |
|---|---|---|---|---|---|---|---|---|---|---|
| Metrics | South | North | South | North | South | North | South | North | South | North |
| TP | 91 | 411 | 47 | 238 | 186 | 156 | 253 | 164 | 199 | 175 |
| FP | 0 | 36 | 4 | 30 | 9 | 0 | 17 | 0 | 0 | 5 |
| FN | 160 | 195 | 3 | 19 | 25 | 36 | 22 | 102 | 1 | 45 |
| Precision | 1.000 | 0.919 | 0.922 | 0.888 | 0.954 | 1.000 | 0.937 | 1.000 | 1.000 | 0.972 |
| Recall | 0.363 | 0.678 | 0.940 | 0.926 | 0.882 | 0.813 | 0.920 | 0.617 | 0.995 | 0.795 |
| F1 | 0.532 | 0.781 | 0.931 | 0.907 | 0.916 | 0.897 | 0.928 | 0.763 | 0.997 | 0.875 |
| | *50ft–90°* | | *100ft–90°* | | *200ft–90°* | | *300ft–90°* | | *400ft–90°* | |
| Metrics | South | North | South | North | South | North | South | North | South | North |
| TP | 97 | 29 | 64 | 54 | 59 | 85 | 192 | 116 | 112 | 154 |
| FP | 31 | 8 | 3 | 20 | 4 | 11 | 1 | 3 | 1 | 46 |
| FN | 17 | 0 | 10 | 10 | 2 | 9 | 10 | 16 | 63 | 29 |
| Precision | 0.758 | 0.784 | 0.955 | 0.730 | 0.937 | 0.885 | 0.995 | 0.975 | 0.991 | 0.770 |
| Recall | 0.851 | 1.000 | 0.865 | 0.844 | 0.967 | 0.904 | 0.950 | 0.879 | 0.640 | 0.842 |
| F1 | 0.802 | 0.879 | 0.908 | 0.783 | 0.952 | 0.895 | 0.972 | 0.924 | 0.778 | 0.804 |
| | *50ft–135°* | | *100ft–135°* | | *200ft–135°* | | *300ft–135°* | | *400ft–135°* | |
| Metrics | South | North | South | North | South | North | South | North | South | North |
| TP | 91 | 306 | 212 | 59 | 377 | 72 | 148 | 266 | 453 | 187 |
| FP | 4 | 10 | 3 | 2 | 19 | 1 | 0 | 3 | 12 | 2 |
| FN | 1 | 149 | 3 | 3 | 1 | 1 | 0 | 6 | 7 | 1 |
| Precision | 0.958 | 0.968 | 0.986 | 0.967 | 0.952 | 0.986 | 1.000 | 0.989 | 0.974 | 0.989 |
| Recall | 0.989 | 0.673 | 0.986 | 0.952 | 0.997 | 0.986 | 1.000 | 0.978 | 0.985 | 0.995 |
| F1 | 0.973 | 0.794 | 0.986 | 0.959 | 0.974 | 0.986 | 1.000 | 0.983 | 0.979 | 0.992 |

A good detector is one having a good balance between precision and recall, which can be often represented by F1 scores. F1 scores of different scenarios (azimuth angle and height combination) were plotted by road direction and video band (GRB and infrared).

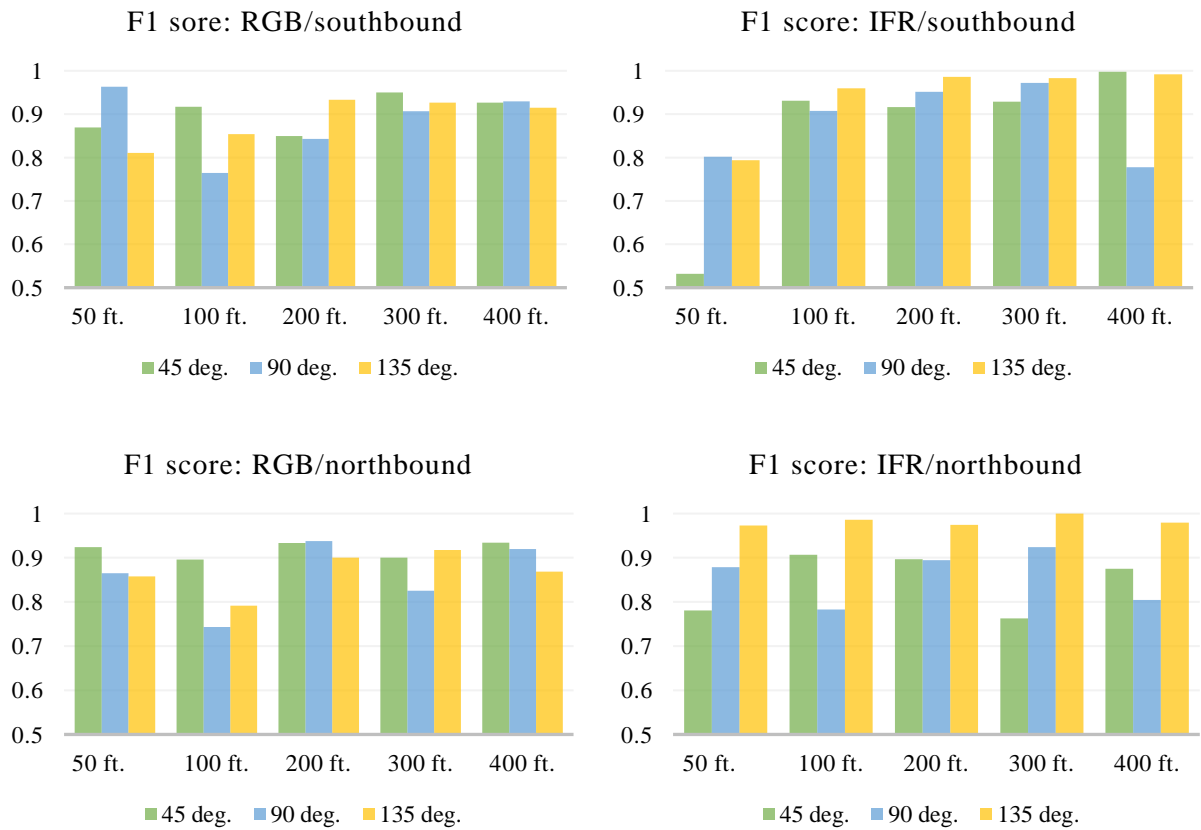

**Figure 7 F1 scores of different scenarios by road direction and video band.**

The performance metrics statistics show overall that the background subtraction-based method applied in this study can achieve good detection performance on RGB images with most F1 scores around 0.9. A higher height (above 200 ft) tends to have more consistent and better performance for different azimuth angles. The scenario of azimuth angle $\psi = 90°$ at 50 ft had a very high F1 score on southbound, which is contrary to expectations, as the algorithm was expected to have difficulty distinguishing vehicles next to each other at an azimuth angle $\psi = 90°$ because vehicles in the front blocking vehicles in the back are more likely to occur than an azimuth angle of $\psi = 45°$ or $\psi = 135°$. After examination of the data, it was found that barely any vehicles were next to each other in the detection time period on southbound. The F1 score would be much lower than the value obtained in the experiment if it was under denser traffic. The blockage issue at a lower height will be solved as the height increases; this is more obvious at an azimuth angle of $\psi = 90°$. **Figure 8** shows an upward trend of recall at the azimuth angle of $\psi = 90°$, as height increased in both directions (except at 50 ft on southbound, as noted). This increase in recall made contributions to F1 score improvement at higher heights. As the height reached over 200 ft, the azimuth angle of $\psi = 90°$ could catch up and even outperform other azimuth angles in recall and F1 scores in some scenarios.

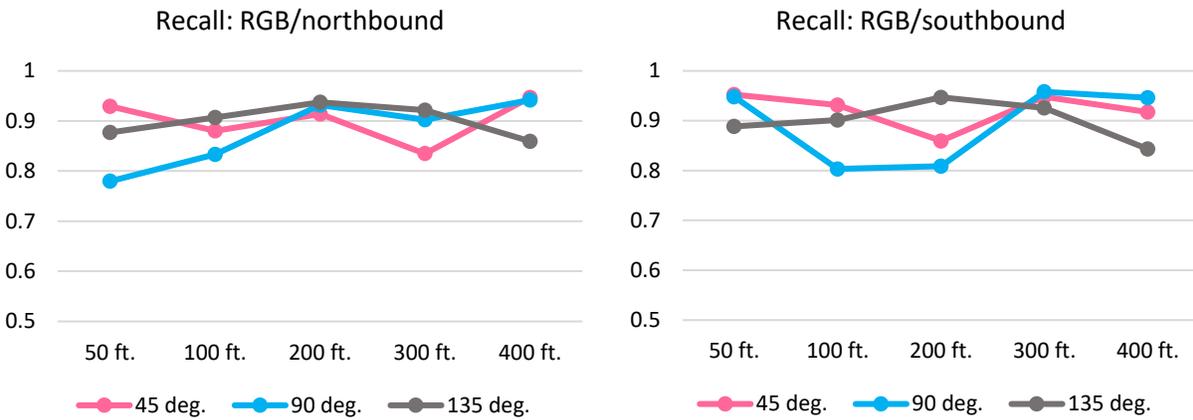

**Figure 8 Recall of northbound traffic on RGB images.**

Compared to RGB images, the results of F1 scores on infrared images show more variation from different azimuth angles. Although the performance seems better with higher heights for southbound, the higher heights in northbound saw a decrease in F1 score for azimuth angles of 45° and 90°. This is because infrared images are more sensitive to noise than RGB images when applying background subtraction-based methods. **Figure 9** shows an example of the detection results when the detection algorithm was run with the same parameters in the model. The inset of the frame shows the infrared image; the rest of the frame is RGB. Noise was created from the road median, roadside, lawn, etc., on the infrared image; no noise was generated on the RGB image.

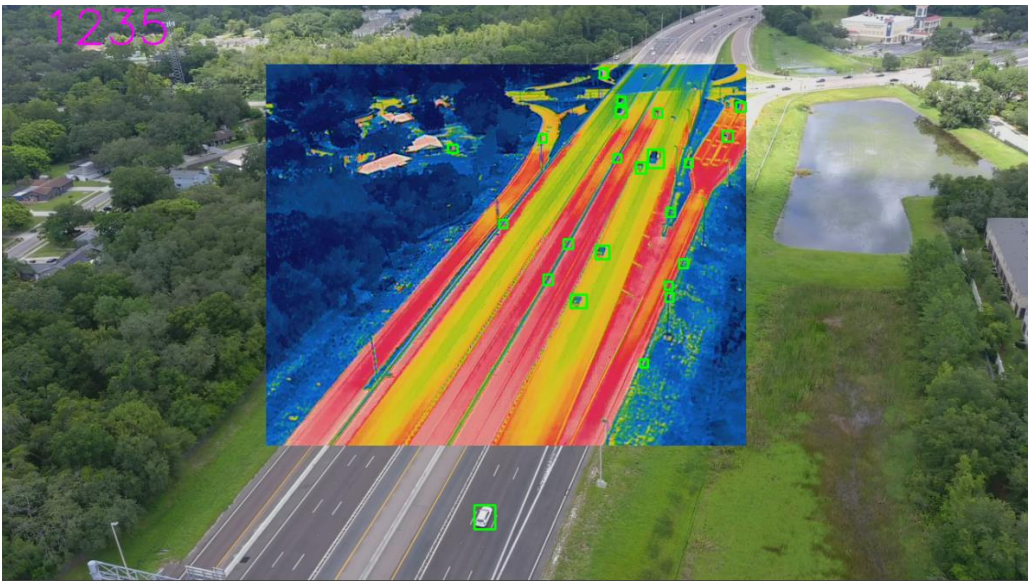

**Figure 9 Detection results of infrared image inset on an RGB image.**

In the Gaussian Mixture-based Background/Foreground Segmentation Algorithm, there is a parameter that sets the threshold on the squared Mahalanobis distance between the pixel and the model. This parameter is used to decide whether a pixel is well-described by the background model. To remove additional noise from infrared images, this parameter needs to be increased to include the noise in the background rather than outputting them as foreground; as a result, some vehicles that should output as foreground would be mistakenly labeled as background, leading to declining recall values (see **Figure** 10). On the other hand, precision may be impacted when achieving a higher recall by decreasing the threshold. In the end, F1 scores are likely to be impacted by a reduction in recall or precision due to noise. The results show that the F1 score on southbound was significantly impacted by noise for scenarios of azimuth angle $\psi = 45°$ on 50 ft and azimuth angle $\psi = 90°$ on 400 ft; on northbound, azimuth angle $\psi = 45°$ on 300 ft and azimuth angle $\psi = 90°$ on 400 ft were also impacted.

Recall: 45 deg./northbound

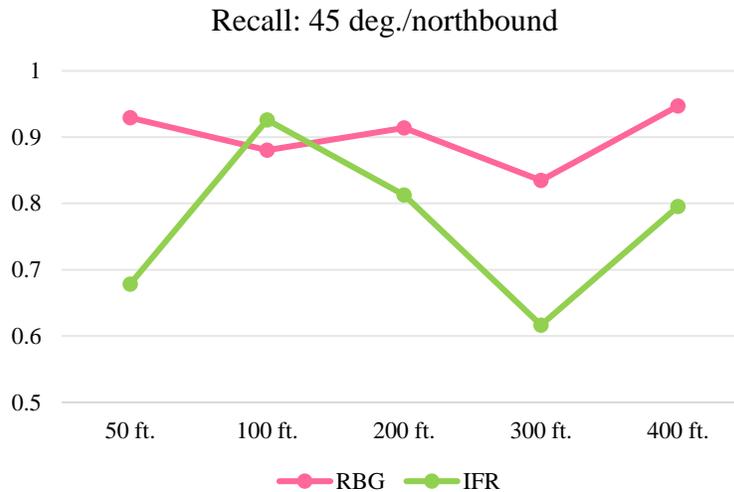

**Figure 10 Recall of azimuth angle $\psi = 45°$ on RGB and infrared images.**

**DISCUSSION**

According to the literature review, practitioners lack a tool that can provide real-time incident detection without violating privacy protection. Therefore, this study explored real-time vehicle detection algorithms using both visual and infrared cameras. The application of UAS with different sensing technologies for obtaining real-time traffic operational information of freeways was explored. Video data in both visual and infrared bands were collected along interstate highways in the Tampa area, and experiments were conducted to quantify the performance of a real-time background subtraction-based method in vehicle detection from a stationary camera (drone hovering at a fixed station) under free-flow conditions. Finally, the relationship between experimental parameters and performance metrics was analyzed.

The experiment outcomes show that, overall, the background subtraction-based method applied in this study can achieve good detection performance on RGB images, with most F1 scores around 0.9. A higher height (above 200 ft) tends to have more consistent and better performance for different azimuth angles. Compared to RGB images, the performance of infrared images had more variations from different azimuth angles, with only some F1 scores better than or comparable to RGB images. This is because infrared images are more sensitive to noise, which affects precision. To reduce noise, the threshold on the squared Mahalanobis distance could be increased to include those noise to the background, which will improve the precision but will inevitably impact recall. For background subtraction-based methods, the detection performance of infrared images has the potential to outperform RDB images if the camera is stable and little noise is made. More trial and error efforts need to be conducted to investigate the best way to reduce noise while analyzing infrared images.

As noted, more data have been collected for different combinations of parameters for traffic conditions with and without congestion and with a drone hovering or moving along a freeway. The results shown in this paper cover only a subset of the data; this effort is ongoing, and results and insights will be shared at a later date.

*Tang, H., Post, J., Kourtellis, A., Porter, B., and Zhang, Y.*

**ACKNOWLEDGMENT**

The authors gratefully acknowledge the support provided by the National Center for Congestion (NICR), University Transportation Centers sponsored by the US Department of Transportation through Grant No. 69A3551947136. The contents of this manuscript reflect the views of the authors, who are responsible for the facts and accuracy of the information presented herein. The contents do not necessarily reflect the official views or policies of US Department of Transportation. The authors also want to acknowledge of the support of Sonali Kannaujia and Keerthana Yelchuri from University of South Florida and Luz Gabriela Rivera Perez and Bryan E Ruiz Hernandez from UPRM for obtaining ground truth of vehicle counts.